\documentclass[a4paper,twoside]{article}

\usepackage{epsfig}
\usepackage{subcaption}
\usepackage{calc}
\usepackage{amssymb}
\usepackage{amstext}
\usepackage{amsmath}
\usepackage{amsthm}
\usepackage{multicol}
\usepackage{pslatex}
 \usepackage{natbib}
 \let\bibhang\relax
\usepackage{apalike}
\usepackage{algorithm2e}
\usepackage[bottom]{footmisc}

\usepackage{url}
\usepackage{booktabs}
\usepackage{dirtytalk}
\usepackage{float}
\usepackage{xcolor}

\usepackage{hyperref}
\hypersetup{colorlinks=true, citecolor=black, linkcolor=black, urlcolor=blue}

\usepackage{SCITEPRESS}     % Please add other packages that you may need BEFORE the SCITEPRESS.sty package.

\begin{document}

\title{Lexical Substitution is not Synonym Substitution: \\On the Importance of Producing Contextually Relevant Word Substitutes}

\author{\authorname{Juraj Vladika\orcidAuthor{0000-0002-4941-9166}, Stephen Meisenbacher\orcidAuthor{0000-0001-9230-5001} and Florian Matthes\orcidAuthor{0000-0002-6667-5452}}
 \affiliation{Department of Computer Science, School of Computation, Information and Technology, \\Technical University of Munich, Germany}
 \email{\{juraj.vladika, stephen.meisenbacher, matthes\}@tum.de}
 }


\abstract{Lexical Substitution is the task of replacing a single word in a sentence with a similar one. This should ideally be one that is not necessarily only synonymous, but also fits well into the surrounding context of the target word, while preserving the sentence's grammatical structure. Recent advances in Lexical Substitution have leveraged the masked token prediction task of Pre-trained Language Models to generate replacements for a given word in a sentence. With this technique, we introduce \textsc{ConCat}, a simple augmented approach which utilizes the original sentence to bolster contextual information sent to the model. Compared to existing approaches, it proves to be very effective in guiding the model to make contextually relevant predictions for the target word. Our study includes a quantitative evaluation, measured via sentence similarity and task performance. In addition, we conduct a qualitative human analysis to validate that users prefer the substitutions proposed by our method, as opposed to previous methods. Finally, we test our approach on the prevailing benchmark for Lexical Substitution, CoInCo, revealing potential pitfalls of the benchmark. These insights serve as the foundation for a critical discussion on the way in which Lexical Substitution is evaluated.}

\onecolumn \maketitle \normalsize \setcounter{footnote}{0} \vfill

\section{INTRODUCTION}

Lexical substitution (LS) can be described as the task of replacing a word in a sentence with the most appropriate different word. It is one of the essential linguistic tasks in Natural Language Processing (NLP) and is an integral component of more complex NLP tasks that deal with text rewriting. Some generative tasks where LS is an important part include paraphrasing \citep{fu2019paraphrase},  machine translation \citep{agrawal-carpuat-2019-controlling}, style transfer \citep{helbig-etal-2020-challenges},
%author anonymization \citep{bo-etal-2021-er}, 
defense against adversarial attacks \citep{Zhou2021DefenseAS}, text simplification \citep{vstajner2022lexical}, or private text rewriting \citep{meisenbacher-etal-2024-dp}. 

%While at a first glance lexical substitution might be trivialized as a mere lookup of synonyms in a large thesaurus of words, the task is more nuanced than that. 
More so than just replacing a word in text with the one of the most similar meaning from a thesaurus, we argue that true LS also ideally regards the surrounding context of the target word and tries to produce candidates that best fit the semantic flow of the whole sentence. In tasks dealing with text rewriting, the goal is to preserve the semantic meaning of the sentence, where simply choosing synonyms may be inadequate. 
%For example, in the sentence "\textit{I got ten \textbf{questions} right in the exam}", the synonym replacement could lead to "\textit{I got ten \textbf{inquiries} right in the exam}", while a more contextual substitution would be "\textit{I got ten \textbf{answers} right in the exam}". In this case, it is in fact a word that is an antonym (\textit{answers}) to the target word (\textit{questions}) that is the most appropriate replacement.

%TODO: Move most parts from here to method description section
%The task of lexical substitution has been tackled with various approaches throughout recent decades. Early methods relied on rule-based systems, heuristics, and lookup to thesaurus databases such as WordNet. 
While early LS methods relied on rule-based systems, pre-trained language models (PLMs) like BERT \citep{devlin-etal-2019-bert} have been predominantly used in recent years. The training objective of PLMs with a masked token prediction (masked language modeling, MLM) provides a natural approach to LS. While this mechanism has been utilized in current LS methods, they are prone to overfitting to the target word by predicting pure synonyms.
%or overfitting to the surrounding context by predicting words with completely different meanings. 
For this purpose, we introduce \textsc{ConCat}, a method for English lexical substitution that in a simple way of concatenating the masked sentence with the target sentences provides an improved trade-off between semantics and context.
%by utilizing the idea of next sentence prediction (NSP). This approach is based on the similar idea used for lexical simplification by \citep{qiang2020BERTLS}.

To test the performance of \textsc{ConCat}, we deploy both quantitative and qualitative analysis. We first %perform a numeric evaluation by looking at prediction scores of our model on 
use three standard benchmarks for lexical substitution, namely LS07 \citep{mccarthy-navigli-2007-semeval}, CoInCo \citep{kremer-etal-2014-substitutes}, and Swords \citep{lee-etal-2021-swords}. Even though our model shows satisfying results on these datasets, upon manual inspection of the gold substitutes provided by the annotators, we noticed a number of problems and inconsistencies. This leads to a critical discussion of the way LS is currently evaluated in the NLP community. To address this gap, we perform a qualitative survey in form of a questionnaire, where users were invited to choose their preferred lexical substitutes in four different settings. 
%The users were shown example CoInCo sentences and provided both with the gold substitutes from the dataset and generated replacements by our model and a competing model. 
The analysis of the results shows that users highly favor substitutes generated by our approach and deem them the most appropriate in given context. While this type of human evaluation is common in generative NLP tasks, to the best of our knowledge there has not been such an evaluation for LS.

Finally, we test to what extent LS preserves the semantic usefulness of text. We achieve this by lexically substituting words in the input sentences of a text classification dataset and observing the effect on a trained classification model. 
%We posit that if the LS approach truly preserves semantic meaning of the sentences, then this should not lead to a big drop in classification performance on the task. We opt for topic classification and the AG News dataset as a representative task and dataset for this experiment. We replaced words in four different frequency settings. 
The results reveal that the model performance stays very close to the original text, especially in the case of \textsc{ConCat} substitutions.  
%Given that lexical substitution of salient words is a common type of an adversarial task in NLP where the goal is to flip the performance of a model, this opens the door to new insights and reconsideration of common techniques in this field as well.

\iffalse
Our contributions are as follows:
\begin{itemize}
    \item We introduce \textsc{ConCat}, a simple and effective lexical substitution method for English that is capable of producing highly contextually fitting replacement words.
    \item We evaluate the method on three standard LS benchmarks: LS07, CoInCo, and Swords. We provide a critical discussion of the benchmarks and their potential shortcomings.
    \item We conduct a qualitative survey where we inquire, for a given sentence from CoInCo and a target word to replace, which words do the survey users prefer among the generated substitutes and original substitutes provided by CoInCo annotators. We show that users highly favor substitutes generated by \textsc{ConCat} and deem them as the most appropriate in given context.
    \item We examine how well \textsc{ConCat} preserves semantic meaning of text by testing the performance of a topic classification model on the original and lexically substituted input text. We show that our approach is in most cases better at preserving the performance of the model than a competing LS approach.
\end{itemize}
\fi

Our contributions are as follows:
\begin{itemize}
    \item We introduce \textsc{ConCat}, a simple lexical substitution method for English, capable of producing highly contextually fitting replacement words.
    \item We evaluate the method on three standard LS benchmarks: LS07, CoInCo, and Swords. We provide a critical discussion of the benchmarks and point out their shortcomings.
    \item We conduct a qualitative survey and show that users highly favor substitutes generated by \textsc{ConCat} and deem them most appropriate contextually, even when compared to gold substitutes.
    \item We examine how well \textsc{ConCat} preserves the semantic meaning of text, showing that          \textsc{ConCat} is in most cases the best at preserving performance.
    \item We make the code for \textsc{ConCat} publicly available at \url{https://github.com/sebischair/ConCat}.
\end{itemize}

\begin{figure}[ht!]
    \centering
    \includegraphics[width=0.99\linewidth]{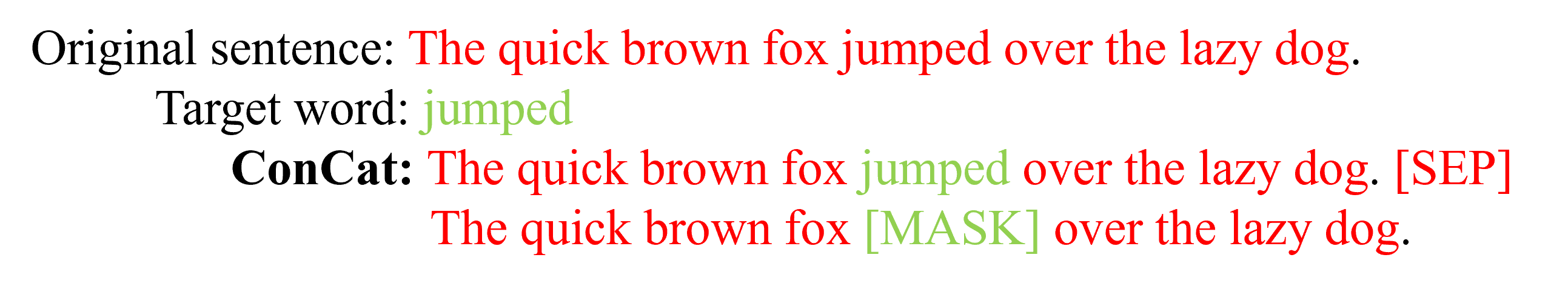}
    \caption{\textsc{ConCat}: a simple and intuitive method for contextually relevant Lexical Substitution.}
    \label{fig:example}
\end{figure}

\section{RELATED WORK}
\label{sec:related}

Lexical substitution was formally defined by \citet{mccarthy2002lexical}.
%and later featured as a shared task in SemEval 2007 \citep{mccarthy2009english}. 
Early approaches used rule-based heuristics and synonym lookup in word thesauri \citep{hassan2007unt}.
%or distributional sparse-vector representations \citep{dinu-lapata-2010-measuring}. 
With the advent of word embedding methods,
%like word2vec \citep{DBLP:journals/corr/abs-1301-3781}, 
LS approaches began representing the target word and its context with dense vectors and ranking the candidates with vector-similarity metrics \citep{roller2016pic}. Most recent approaches utilize pre-trained language models (PLMs). 

Word representations learned by PLMs are highly contextual and finding substitutes that fit the word's surrounding context was significantly improved. Since the prediction of substitutes in this manner can be highly biased towards the target word, \citet{zhou-etal-2019-bert} apply a dropout mechanism by resetting some dimensions of the target word's embedding and then leverage the model to predict substitutes using this perturbed embedding. Later methods by \citet{michalopoulos-etal-2022-lexsubcon} and \citet{seneviratne-etal-2022-cilex} combine the PLM word embeddings with a gloss value of target word's synonyms returned by WordNet
%, enhanced with an auxiliary gloss regularizing component from \citep{lin-etal-2022-improving},
to guide the vector-space exploration towards similar words. \citet{qiang-etal-2023-parals} generate substitutes with paraphrase modeling and improved decoding.

Unlike the approaches that use WordNet for embedding perturbation, our approach includes querying WordNet for a rule-based filtering of unsuitable words.
%(like antonyms and related forms of the target word). 
To the best of our knowledge, we are the first to utilize the idea of sentence concatenation for improved LS and the first to do a qualitative analysis of generated substitutes with human evaluation.%, as is common practice in other NLG tasks. 
%Instead of focusing just on benchmark results, we aim to conduct a qualitative investigation into the generated substitutes, in order to assess their usefulness in real-world text rewriting use cases.

% Alasca \citep{lacerra2021alasca}, LexSubCon \citep{michalopoulos-etal-2022-lexsubcon}, Cilex \citep{seneviratne-etal-2022-cilex}. %Gloss loss \citep{lin-etal-2022-improving}

\section{METHODOLOGY}
\label{sec:method}
\subsection{The \textsc{ConCat} Approach}
%As with most other NLP tasks, pre-trained language models have been used in recent years as the backbone of LS systems. 
Language models are powerful predictors of words that contextually fit a sentence, due to their MLM learning objective of predicting a missing word in a sentence, performed over massive training corpora. An intuitive approach to LS is simply to replace the target word with a [MASK] token and let the PLM predict top candidates. Another approach would be to keep the original text intact to predict substitutes. 
From our observation, the first approach tends to produce contextually relevant words that can be semantically distant from the original word, while the second approach overfits to the target word and predicts solely its different inflectional forms or synonyms, thus lacking creativity. To bridge this gap, our method combines these two approaches -- the masked sentence (with the target word masked) is concatenated with the original sentence, separated by a separator token. 
%This approach leverages both the MLM and NSP training objectives of BERT-style models to predict both semantically consistent and contextually fitting word substitutes. 
This was inspired by a similar concatenation method for lexical simplification \citep{qiang2020BERTLS}.

\textsc{ConCat}, our proposed simple and intuitive LS method, is demonstrated in Figure \ref{fig:example}.
%The original sentence has the target word masked out with the predefined [MASK] token, which was learned by the model in its pre-training MLM objective. This masked sentences is concatenated to the original sentence and separated with the predefined [SEP] token.
%another token that the model associates with its training process. 
This approach combines the best of both worlds -- it increases creativity by forcing the model to predict an empty [MASK] token but also makes the model aware of the original word by including it in the next sentence. After experimenting with multiple base models such as BERT \citep{devlin-etal-2019-bert} and XL-Net \citep{yang2019xlnet}, we opted for RoBERTa \citep{DBLP:journals/corr/abs-1907-11692}, which seemed to produce the most fitting substitutes. Additionally, its tokenizer uses full words as tokens, facilitating direct word substitution. We use the \textsc{roberta-base} model from HuggingFace since it is less resource-intensive than the large variant, without a significant performance drop.

Despite the impressive performance, there was still a considerable number of instances where our approach fell short, resorting to predicting words which are antonyms of the target word (which could fit the context but not the semantics of the sentence) or grammatical variations of the target word (which breaks the grammatical correctness of the sentence). To overcome this obstacle, we additionally deploy checks based on WordNet and filter out inadequate words, creating a hybrid approach.
In particular, we use Wordnet to obtain lists of synonyms, antonyms, hypernyms, hyponyms, meronyms, holonyms, and then manually filter out these from the top $k$ generated candidates of the masked target word.

\begin{table*}[htbp]
\caption{Benchmark results for both the LS07 and CoInCo tasks. Metrics were calculated using the approach described by \citep{mccarthy-navigli-2007-semeval}. Highest scores are \textbf{bolded}.}
\resizebox{\linewidth}{!}{
\begin{tabular}{lcccccc|cccccc}
\toprule
\multicolumn{1}{r|}{\textbf{Dataset:}} & \multicolumn{6}{c|}{\textbf{LS07}} & \multicolumn{6}{c}{\textbf{CoInCo}} \\
\multicolumn{1}{r|}{\textbf{Model:}} & \multicolumn{1}{l}{best} & \multicolumn{1}{l}{best-mode} & \multicolumn{1}{l}{oot} & \multicolumn{1}{l}{oot-mode} & \multicolumn{1}{l}{P@1} & \multicolumn{1}{l|}{P@3} & \multicolumn{1}{l}{best} & \multicolumn{1}{l}{best-mode} & \multicolumn{1}{l}{oot} & \multicolumn{1}{l}{oot-mode} & \multicolumn{1}{l}{P@1} & \multicolumn{1}{l}{P@3} \\
\midrule
\midrule
\multicolumn{1}{r|}{Dropout} & 20.3 & 34.2 & 55.4 & 68.4 & 51.1 & -- & 14.5 & 33.9 & 45.9 & 69.9 & 56.3 & -- \\
\multicolumn{1}{r|}{Dropout*} & 2.37 & 12.81 & 23.73 & 32.02 & 16.67 & 26.67 & 3.44 & 10.57 & 18.91 & 23.59 & 23.54 & 40.82 \\
\multicolumn{1}{r|}{\textsc{ConCat}} & \textbf{3.52} & \textbf{16.26} & \textbf{35.11} & \textbf{46.31} & \textbf{21.67} & \textbf{40.33} & \textbf{4.81} & \textbf{15.63} & \textbf{26.07} & \textbf{34.47} & \textbf{34.66} & \textbf{53.81} \\
\bottomrule
\end{tabular}
}
\label{tab:eval}
\end{table*}

\begin{table*}[htbp]
\caption{Benchmark results for the Swords task. Metrics were calculated and are presented as in Table \ref{tab:eval}.}
\resizebox{\linewidth}{!}{
\begin{tabular}{lcccccc|cccccc}
\toprule
\multicolumn{1}{r|}{\textbf{Dataset:}} & \multicolumn{6}{c|}{\textbf{Swords 1}} & \multicolumn{6}{c}{\textbf{Swords 5}} \\
\multicolumn{1}{r|}{\textbf{Model:}} & \multicolumn{1}{l}{best} & \multicolumn{1}{l}{best-mode} & \multicolumn{1}{l}{oot} & \multicolumn{1}{l}{oot-mode} & \multicolumn{1}{l}{P@1} & \multicolumn{1}{l|}{P@3} & \multicolumn{1}{l}{best} & \multicolumn{1}{l}{best-mode} & \multicolumn{1}{l}{oot} & \multicolumn{1}{l}{oot-mode} & \multicolumn{1}{l}{P@1} & \multicolumn{1}{l}{P@3} \\
\midrule
\midrule
\multicolumn{1}{r|}{Dropout*} & 1.34 & \textbf{3.39} & \textbf{13.44} & \textbf{18.64} & 22.43 & 20.27 & 2.19 & \textbf{8.33} & \textbf{21.92} & \textbf{27.27} & 40.54 & 32.70 \\
\multicolumn{1}{r|}{\textsc{ConCat}} & \textbf{2.43} & \textbf{3.39} & 13.17 & 13.56 & \textbf{35.95} & \textbf{31.08} & \textbf{4.02} & \textbf{8.33} & 20.76 & 25.00 & \textbf{48.38} & \textbf{42.43} \\
\bottomrule
\end{tabular}
}
\label{tab:eval2}
\end{table*}

\subsection{Evaluation}
To evaluate our new method, we conduct a multi-headed evaluation consisting of three parts: (1) evaluation on the LS07, CoInCo, and Swords benchmarks, (2) evaluation and analysis of LS on a classification task, and (3) a qualitative evaluation led by surveys.

\paragraph{LS Benchmarking}
We begin our evaluation with measuring our method's performance on popular LS bechmarks: LS07 \citep{mccarthy-navigli-2007-semeval}, CoInCo \citep{kremer-etal-2014-substitutes}, and Swords \citep{lee-etal-2021-swords}. In all of them the goal is to provide substitutes for given target words in the provided sentences. Note that in the case of Swords, we employ two versions of the provided gold labels: (1) \textit{Swords 1}, which use all gold labels where at least one annotator voted it to be suitable, and (2) \textit{Swords 5}, where at least 50\% of annotators agreed upon a substitute.

As metrics, we employ the four LS metrics from SemEval 2007.
%, namely \textit{best}, \textit{best-mode}, \textit{oot}, and \textit{oot-mode}. 
We briefly introduce them, but refer the reader to the original task report for more details.

\begin{itemize}
\small
    \itemsep 0em
    \item \textit{best}: evaluates the quality of the best prediction, by scoring the existence of the gold top substitute, weighted by the order.
    \item \textit{best-mode}: evaluates if the system's best prediction appears in the \textit{mode} of the gold labels.
    \item \textit{oot}: evaluates the coverage of the gold substitutes in the system predictions, i.e., the percentage of gold substitutes appearing in the system's top-10 predictions.
    \item \textit{oot-mode}: evaluates the \% of times the mode of the gold labels appears in the system's top-10 predictions.
\end{itemize}

In addition to these metrics, we also employ P@1 and P@3. It should be noted that for the entire benchmark, we limit the system predictions to 10 responses, so as to not skew the \textit{best} scoring.

\paragraph{LS Task Performance}
The second step of our evaluation includes measuring performance on a NLP task after performing LS on the original dataset. For this, we select the AG News dataset \citep{Zhang2015CharacterlevelCN}, which presents a Multi-Class Topic Classification task. We take a random sample of 10,000 rows for our evaluation.
For the task, 
%we envision four experimental settings, governed by the percentage of tokens in the original dataset that are replaced. In particular, 
we select four settings $Subst. \% \in \{0.25, 0.5, 0.75, 1.0\}$, where the percentage represents the randomly selected percentage of tokens in the dataset that are replaced via LS. We use our proposed method (\textsc{ConCat}), as well as the method proposed by \citet{zhou-etal-2019-bert}. 

To measure performance (accuracy) in each of these settings, we use each resulting dataset to train a LSTM model \citep{hochreiter1997long} using keras. As an input layer, we use GloVe embeddings \citep{pennington-etal-2014-glove}. For training, we set the batch size to 64 and train for a maximum of 30 epochs with early stopping. Accuracy is captured from the best resulting model, and the training process is run five separate times to obtain average accuracy.

\begin{table}[htbp]
\small
\caption{Comparing CoInCo gold substitutions to ours. Top-1 took the top candidate from each substitution set and replaced the target. Random-1 picked a word from the substitution set at random (same index for gold and ours).}
\centering
\resizebox{\linewidth}{!}{
\begin{tabular}{rclllll}
\toprule
\textbf{Task:} & \multicolumn{6}{c}{\textbf{CoInCo (Cosine Similarity)}} \\
\midrule
\textbf{Substitution:} & \multicolumn{3}{c}{Top-1} & \multicolumn{3}{c}{Random-1} \\
\textbf{Model:} & \multicolumn{1}{l}{Mini} & DR & MPN & Mini & DR & MPN \\
\midrule
\midrule
\multicolumn{1}{l|}{CoInCo Gold} & \multicolumn{1}{l}{76.15} & 69.67 & \multicolumn{1}{l|}{72.08} & 76.26 & 69.88 & 72.27 \\ \cline{2-7} 
\multicolumn{1}{r|}{Avg:} & \multicolumn{3}{c|}{72.63} & \multicolumn{3}{c}{\textbf{72.80}} \\
\midrule
\multicolumn{1}{l|}{\textsc{ConCat}} & \multicolumn{1}{l}{78.00} & 70.24 & \multicolumn{1}{l|}{73.10} & 76.53 & 69.44 & 71.81 \\ \cline{2-7} 
\multicolumn{1}{r|}{Avg:} & \multicolumn{3}{c|}{\textbf{73.77}} & \multicolumn{3}{c}{72.59} \\
\bottomrule
\end{tabular}
}
\label{tab:CoInCo_cos}
\end{table}

\paragraph{Perplexity Test}
\textit{Perplexity} is a common method for the evaluation of generative LMs \citep{chen1998evaluation}. Essentially, perplexity measures how \say{uncertain} a LM is in making a next token prediction. A lower perplexity, therefore, implies that an LM is less \say{surprised} by the context given to the model. Perplexity can be used to assess text complexity of a sentence \citep{vladika-etal-2022-tum}. We aim to leverage this metric to evaluate how well a word substitute fits into its context sentence. In particular, we envision two settings:
\begin{itemize}
    \footnotesize
    \itemsep 0em
    \item \textbf{Top-10}: all top-10 generated substitutes are replaced into the sentence, and the average perplexity of these sentences is taken.
    \item \textbf{Top-Match}: only to the top $k$ substitutes are measured, where $k$ denotes the number of gold substitutes.
\end{itemize}
These results are compared against the perplexity of the original context sentences (\textit{baseline}), and the average perplexity of the sentences replaced with the gold substitutes (\textit{gold}). For perplexity, we use the GPT2 model \citep{radford2019language}. 

\paragraph{Qualitative Analysis}
In the final stage of our evaluation, we administer a survey via Google Forms, with the goal of evaluating user preference of  different LS methods' predictions. To accomplish this, we divide the survey into four tasks:

\begin{enumerate}
\small
    \itemsep 0em
    \item \textbf{Single Word Replacement (SWR)}: a sentence is presented with the target word bolded, and the user is asked to select from a list of single substitutes.
    \item \textbf{Single Word Replacement, Masked (SWR-M)}: a sentence is shown with a placeholder instead of the target word to be replaced, and the user is asked to select, from a list a substitutes, which replacement is most suitable.
    \item \textbf{Set Replacement (SR)}: setup similar to SWR, but the list of options includes sets of three replacements, and the user must select which \textit{set} is most suitable.
    \item \textbf{Set Replacement, Masked (SR-M)}: like SR, but the target word is once again masked.
\end{enumerate}

For the survey questions, we select 60 random entries from CoInCo, 15 for each task. As answer options, we present the top 1 or 3 gold substitutes from CoInCo, the top 1 or 3 using the method of \citet{zhou-etal-2019-bert}, and the top 1 or 3 from \textsc{ConCat}.

\section{EXPERIMENT RESULTS}
\label{sec:results}
%In this section, we introduce the results of our three-step evaluation.

\paragraph{Benchmark}
The results for the evaluation on the three benchmark datasets are in Tables \ref{tab:eval} and \ref{tab:eval2}. For all tasks, we include 6 metrics, outlined in the previous section. These metrics are placed in juxtaposition to \citet{zhou-etal-2019-bert} for comparison. The original scores of \citet{zhou-etal-2019-bert} could not be replicated due to unavailability of the original code. Therefore, we reimplemented their method and include our replicated score as well, marked by an asterisk (*). 
%This score gives a more realistic picture.

As an added point of comparison, we compute cosine similarity scores for our method's replacements, and the gold CoInCo substitutes. To compute these similarity scores, we utilize three Sentence Transformer models \citep{reimers-2019-sentence-bert}: \textsc{all-MiniLM-L12-v2} (Mini), \textsc{all-distilroberta-v1} (DR), and \textsc{all-mpnet-base-v2} (MPN).  In addition, we compute the scores in two settings: (1) \textbf{Top-1}, where the target word is replaced with the top annotator response / system prediction, and (2) \textbf{Random-1}, where the target is replaced with a randomly chosen substitute. These results are in Table \ref{tab:CoInCo_cos}.

\begin{table}[htbp]
\small
\caption{Accuracy scores for AG News. \% Subst. denotes the percentage of tokens in the dataset per that were substituted. Scores represent an average of five evaluated models.}
\centering
\resizebox{\linewidth}{!}{
\begin{tabular}{l|cccc}
\toprule
\multicolumn{1}{r|}{Task:} & \multicolumn{4}{c}{\textbf{AG News} (baseline = 88.41 $\pm$ 0.40)} \\
\midrule
\multicolumn{1}{r|}{\% Subst.} & 25\% & 50\% & 75\% & 100\% \\
\midrule
\midrule
Dropout & \textbf{87.52} $\pm$ 0.13 & 85.91 $\pm$ 0.54 & 83.37 $\pm$ 0.36 & 81.37 $\pm$ 0.37 \\
\textsc{ConCat} & 87.41 $\pm$ 0.48 & \textbf{86.18} $\pm$ 0.44 & \textbf{84.48} $\pm$ 0.50 & \textbf{83.04} $\pm$ 0.30 \\
\bottomrule
\end{tabular}
}
\label{tab:acc}
\end{table}

\begin{table*}[htbp]
\small
\caption{Average cosine similarity scores between original dataset and datasets with a percentage of token replacements. SentenceTransformers models used: \textsc{all-minilm-l12-v2} (Mini), \textsc{all-distilroberta-v1} (DR), and \textsc{all-mpnet-base-v2} (MPN). Highest average score per \% Subst. is \textbf{bolded.}}
\centering
\resizebox{0.85\linewidth}{!}{
\begin{tabular}{l|lll|lll|lll|lll}
\toprule
\multicolumn{1}{r|}{\textbf{Task:}} & \multicolumn{12}{c}{\textbf{AG News (Cosine Similarity)}} \\
\midrule
\multicolumn{1}{r|}{\textbf{\% Subst.}} & \multicolumn{3}{c}{25\%} & \multicolumn{3}{c}{50\%} & \multicolumn{3}{c}{75\%} & \multicolumn{3}{c}{100\%} \\
\multicolumn{1}{r|}{\textbf{Model:}} & Mini & DR & MPN & Mini & DR & MPN & Mini & DR & MPN & Mini & DR & MPN \\
\midrule
\midrule
\multicolumn{1}{l|}{Dropout} & \multicolumn{1}{l}{91.23} & 91.00 & \multicolumn{1}{l|}{91.04} & 80.73 & 81.19 & \multicolumn{1}{l|}{80.59} & 69.77 & 71.36 & \multicolumn{1}{l|}{69.82} & 55.51 & 59.37 & 55.76 \\ \cline{2-13} 
\multicolumn{1}{r|}{Avg:} & \multicolumn{3}{c|}{91.09} & \multicolumn{3}{c|}{80.84} & \multicolumn{3}{c|}{70.32} & \multicolumn{3}{c}{56.88} \\
\midrule
\multicolumn{1}{l|}{\textsc{ConCat}} & \multicolumn{1}{l}{91.03} & 91.38 & \multicolumn{1}{l|}{91.34} & 81.05 & 82.07 & \multicolumn{1}{l|}{81.20} & 70.45 & 73.22 & \multicolumn{1}{l|}{71.00} & 57.76 & 62.91 & 58.72 \\ \cline{2-13} 
\multicolumn{1}{r|}{Avg:} & \multicolumn{3}{c|}{\textbf{91.34}} & \multicolumn{3}{c|}{\textbf{81.44}} & \multicolumn{3}{c|}{\textbf{71.56}} & \multicolumn{3}{c}{\textbf{59.80}} \\
\bottomrule
\end{tabular}
}
\label{tab:ag_cos}
\end{table*}

\begin{table*}[htbp]
\small
\caption{Survey responses from 21 respondents. For each of the four tasks, we include the number of responses that preferred a given method's replacement, as well as the corresponding percentage for this task (of 21*15=315 responses per task). Note that in some cases, the total percentage may exceed 100. This occurs when more than one method outputs the same prediction -- both methods would then be counted if selected by a respondent.}
\centering
\begin{tabular}{l|llllllllrr}
\toprule
 & \multicolumn{10}{c}{\textbf{Survey Responses (21 respondents, 60 questions)}} \\
 & \multicolumn{2}{c}{SWR} & \multicolumn{2}{c}{SWR-M} & \multicolumn{2}{c}{SR} & \multicolumn{2}{c}{SR-M} & \multicolumn{2}{r}{\textbf{TOTAL}} \\
\midrule
\midrule
Gold & 80 & \multicolumn{1}{l|}{25.40\%} & \textbf{129} & \multicolumn{1}{l|}{\textbf{40.95\%}} & 119 & \multicolumn{1}{l|}{37.78\%} & 88 & \multicolumn{1}{l|}{27.93\%} & 416 & 33.02\% \\
Dropout & 141 & \multicolumn{1}{l|}{44.76\%} & 114 & \multicolumn{1}{l|}{36.19\%} & 89 & \multicolumn{1}{l|}{28.25\%} & 95 & \multicolumn{1}{l|}{30.16\%} & 439 & 34.84\% \\
ConCat & \textbf{195} & \multicolumn{1}{l|}{\textbf{61.90\%}} & 120 & \multicolumn{1}{l|}{38.40\%} & \textbf{125} & \multicolumn{1}{l|}{\textbf{39.68\%}} & \textbf{112} & \multicolumn{1}{l|}{\textbf{35.56\%}} & \textbf{552} & \textbf{43.81\%} \\
\bottomrule
\end{tabular}
\label{tab:survey}
\end{table*}

\paragraph{AG News}
In Table \ref{tab:acc}, we present the results of our task-based evaluation. Accuracy scores for each subtitution setting (percentage of tokens replaced) are given, once again compared against \citet{zhou-etal-2019-bert}. These results are also plotted in Figure \ref{fig:acc_ag}.

Similarly to benchmark analysis, we employ cosine similarity to illustrate the effect of LS methods on the overall semantics of the underlying AG News dataset. The scores are calculated for each of the substitution settings, and scores from three embedding models are averaged. The results are given in Table \ref{tab:ag_cos}. The results of the perplexity test are found in Table \ref{tab:pp}.

\paragraph{Survey}
The aggregate results from our survey are illustrated in Table \ref{tab:survey}. We received 21 respondents, all who were close colleagues with fully proficient or native levels of English. We separate the results by task, as well as provide total counts.

\section{DISCUSSION}
\label{sec:dicuss}
We now discuss our results in detail, and extract interesting insights from our analysis.

\begin{figure}[htpb]
    \centering
    \includegraphics[scale=0.39]{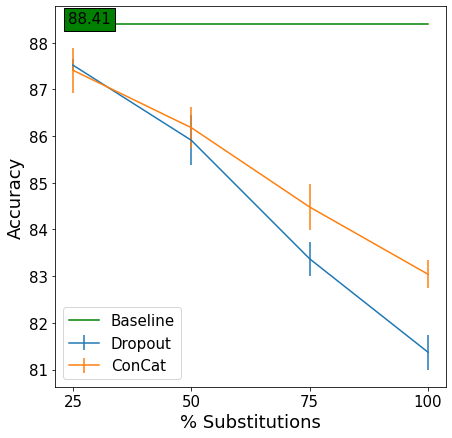}
    \caption{Accuracy scores for LS task performance. As token replacements increase, task performance decreases, but at different rates. Baseline performance shown in green.}
    \label{fig:acc_ag}
\end{figure}

\paragraph{LS vs. Utility}
The task-based evaluation revealed that performing LS on datasets to be fed to downstream tasks does indeed have an effect on model performance. As illustrated by Figure \ref{fig:acc_ag}, a clear degradation in utility occurs as a higher percentage of the AG News dataset is replaced by LS. Interestingly, our method \say{slows} this degradation rate down. 
%We did not attempt replacing \textit{beyond} 100\%, i.e., running LS on the already replaced dataset -- this could prove to provide further insights. Also n
Not covered is \textit{less-than-25\% }replacement, where one could study the intersection of LS and model robustness.

\paragraph{Context Improves Metrics}
The results of applying our method to LS07 and CoInCo clearly show the effects of including contextual information in the LS process. In comparison to \citet{zhou-etal-2019-bert}, whose method does not include such context, our method shows improvement on all metrics. Particularly in the metrics that evaluate \textit{coverage} of the gold substitutes in the system predictions, our method displays considerable improvements. We pose that especially in these cases where coverage is tested, the benefits of contextual information are evident, namely in producing lexically \textit{and} semantically suitable replacements.

\paragraph{Preserving Semantics and Reducing Perplexity}
In evaluating the preservation of semantic similarity, we empirically show that LS in general leads to a loss in information, measured by cosine similarity. 
%Our cosine similarity analysis shows that word replacements inevitably lead to lower scores. 
While this is certainly plausible, the extent to which semantic similarity is affected is shown in Table \ref{tab:CoInCo_cos}, as well in additional task-based experiments.
%whose results are given in Appendix \ref{sec:appendix}.

\begin{table}[htbp]
\caption{Perplexity Results. 
%In all cases, the measured perplexity of the context sentences with lexical substitutes is lower with our replacements than with the gold substitutes. 
Note that for Swords, we use all provided gold labels (\textit{Swords 1}). T-M: Top-Match.}
\centering
\resizebox{\linewidth}{!}{
\begin{tabular}{r|cc|cc|cc}
\toprule
Dataset: &
  \multicolumn{2}{c|}{\textbf{LS07}} &
  \multicolumn{2}{c|}{\textbf{CoInCo}} &
  \multicolumn{2}{c|}{\textbf{Swords}} \\ 
 &
  \multicolumn{1}{l}{Top-10} &
  \multicolumn{1}{l|}{T-M} &
  \multicolumn{1}{l}{Top-10} &
  \multicolumn{1}{l|}{T-M} &
  \multicolumn{1}{l}{Top-10} &
  \multicolumn{1}{l|}{T-M}  \\ \hline
Baseline &
  \multicolumn{2}{c|}{228.62} &
  \multicolumn{2}{c|}{258.19} &
  \multicolumn{2}{c|}{66.78} \\ \hline
Gold &
  \multicolumn{2}{c|}{292.14} &
  \multicolumn{2}{c|}{232.05} &
  \multicolumn{2}{c|}{83.06}  \\ \hline \hline
% \textsc{ConCat} (avg.) &
%   -- &
%   -- &
%   -- &
%   -- &
%   -- &
%   -- \\
\multicolumn{1}{l|}{\textsc{ConCat}} &
  256.36 &
  253.80 &
  206.73 &
  203.58 &
  74.69 &
  73.75  \\
\bottomrule
\end{tabular}
}
\label{tab:pp}
\end{table}

At the same time, the results of the perplexity measurements lead to important insights about the nature of LS. In 3 out of 4 cases, using gold substitutes in sentences
%replacing gold substitutes into their corresponding context sentences actually 
leads to \textit{increased} perplexity. This would imply that the substitutes are not as suitable to maintain natural, flowing sentences. Increased perplexity can also be observed from \textsc{ConCat} replacements, yet the effect of our substitutes is not as severe as those from the gold substitutes. In fact, for CoInCo our replacements lead to \textit{lower} perplexity than baseline.

%While the detailed implications of such results require deeper analysis, one can 
We hypothesize that our replacements might be \say{preferred} by LMs in perplexity measurements, as \textsc{ConCat} aims to produce \textit{contextually} meaningful replacements, rather than substitutes that might be synonymous, but not so fitting in a particular context.

% \subsection{Preserving Semantics}
% In evaluating the preservation of semantic similarity, we empirically show that LS in general leads to a loss in information, measured by cosine similarity. Both in the AG News task and the benchmark, our cosine similarity analysis shows that word replacements inevitably lead to lower scores. While this is certainly plausible, the extent to which semantic similarity is affected is demonstrated by the results shown in Table \ref{tab:ag_cos}.

\paragraph{A Critique of CoInCo}
For evaluation of \textsc{ConCat}, CoInCo was chosen as this dataset serves as the current \textit{de facto} standard benchmark for LS tasks \citep{seneviratne-etal-2022-cilex}. A closer study of its entries, however, reveals intriguing findings that speak to possible considerations to be made in the future.

Firstly, the inclusion of single-word and double-word sentences (e.g., \say{\textit{She \textbf{said}.}}), although a challenging task, greatly biases towards dictionary-based substitutions or similar methods, as context is lacking. 
%The use of the provided context sentences could be a potential solution to this issue. 
Along similar lines, some target words do not consist of full words, but rather word pieces, such as \textit{don} (don't). Similarly, some gold substitute are double-word phrases (e.g., \textit{very much} for \textit{enough}), which is impossible to predict using the single-word prediction of PLMs.
%Clarify?
%As a last note considering the provided sentences, 
Finally, some sentences have not been fully cleaned, which may lead to issues on the system end.

More importantly, some top annotator responses contain errors, e.g., \textit{day} $\rightarrow$ \textit{@card@ hour period}, where \textit{@card@} does not make sense. Moreover, some annotator suggestions are questionable, such as:
%-- consider the following case:

\begin{itemize}
\small
    \setlength{\itemindent}{.2in}
    \itemsep 0em
    \item[\underline{Orig.}] Tasha is not the \textbf{whole} of what happened \\ on Vega IV.
    \item[\underline{Gold}]  \textcolor{red}{bulk}; \textcolor{red}{complete}; \textcolor{red}{consummate}; \textcolor{teal}{entirety}; \\ \textcolor{yellow}{sum}; \textcolor{yellow}{total}
\end{itemize}

While the suggested replacements represent true synonyms of the target, some of these are not contextually suitable, such as \textit{bulk} or \textit{consummate}. 

From these qualitative insights, we decided to conduct a manual inspection of our \textsc{ConCat} method on the provided benchmarks.
%focusing on the \textit{weak points} from the perspective of the gold labels in each benchmark. 
Following a similar inspection performed by \citep{seneviratne-etal-2022-cilex}, we analyze our method's performance on the benchmarks with two metrics: (1) \textit{Top-3 coverage} (\textbf{T3C}): how often each of the top-3 model predictions are in \textit{any} of the gold labels (in \%), and (2) \textit{mismatch percentage} (\textbf{MMP}): how often \textit{none} of the top-3 predictions are in the gold set (in \%). The results of this inspection for \textsc{ConCat} and Dropout* are included in Table \ref{tab:inspect}.

\begin{table}[htbp]
\caption{\textbf{T3C} $\uparrow$ and \textbf{MMP} $\downarrow$ scores for the three benchmark datasets.}
\resizebox{\linewidth}{!}{
    \begin{tabular}{|r|cc|cc|cc|cc|}
    \hline
    Dataset: & \multicolumn{2}{c|}{LS07} & \multicolumn{2}{c|}{CoInCo} & \multicolumn{2}{c|}{Swords 1} & \multicolumn{2}{c|}{Swords 5} \\
     & \multicolumn{1}{l}{T3C} & \multicolumn{1}{l|}{MMP} & \multicolumn{1}{l}{T3C} & \multicolumn{1}{l|}{MMP} & \multicolumn{1}{l}{T3C} & \multicolumn{1}{l|}{MMP} & \multicolumn{1}{l}{T3C} & \multicolumn{1}{l|}{MMP} \\ \hline
    Dropout* & 11.5 & 70.3 & 18.6 & 17.8 & 15.8 & 57.0 & 20.6 & 64.9 \\
    ConCat & 18.6 & 51.3 & 27.7 & 11.9 & 25.3 & 47.0 & 31.4 & 53.7 \\ \hline
    \end{tabular}
}
\label{tab:inspect}
\end{table}

\begin{table*}[htbp]
\caption{Examples from the LS07, CoInCo, and Swords benchmarks. Presented are the original sentences with target words \textbf{bolded}, as well as the annotator gold substitutes and the top-3 predictions from \textsc{ConCat}. Text in \textcolor{red}{red} denotes unsuitable replacements, whereas \textsc{ConCat} substitutes in \textcolor{teal}{green} denote good ones. 
%Note that the non-lemmatized \textsc{ConCat} substitutes are presented for illustrative purposes.
}
\tiny 
\centering
%\resizebox{\linewidth}{!}{
    \begin{tabular}{p{0.35\linewidth}|p{0.40\linewidth}|p{0.16\linewidth}}
    \hline
    \textbf{Sentence} & \textbf{Gold Substitutes} & \textbf{ConCat Top-3} \\ \hline
    \textbf{Finally}, this new rule will also have the effect of encouraging existing corporations to produce safer products & lastly, \textcolor{red}{in} & \textcolor{teal}{note}, \textcolor{teal}{thus}, \textcolor{red}{together} \\ \hline
    Treatment of physical problems , particularly chronic ones , is possible as \textbf{well} as psychological therapy . & \textcolor{red}{in}, \textcolor{red}{along}, \textcolor{red}{including} & \textcolor{teal}{much}, \textcolor{red}{also}, \textcolor{teal}{far} \\ \hline \hline
    Tara \textbf{fumed}. Of all the impertinence! & \textcolor{red}{anger}, be steam, be upset, blow up, boil, \textcolor{red}{bristle}, \textcolor{red}{burn}, \textcolor{red}{digest}, \textcolor{red}{flare}, fret, \textcolor{red}{froth}, glower, grumble, \textcolor{red}{howl}, madden, \textcolor{red}{plan}, rage, rant, \textcolor{red}{rave}, \textcolor{red}{ruffle}, scowl, seethe, \textcolor{red}{smolder}, steam, stew, \textcolor{red}{whine}, yell & \textcolor{teal}{raged}, \textcolor{teal}{complained}, \textcolor{red}{argued} \\ \hline
    I have \textbf{sent} Patti a list. For payment, we have to forecast the money two days out. & \textcolor{red}{convey}, deliver, dispatch, forward, mail, \textcolor{red}{relay}, \textcolor{red}{remit}, \textcolor{red}{report} & \textcolor{teal}{given}, \textcolor{teal}{written}, \textcolor{red}{shipped} \\ \hline \hline
    The factory is highly automated and designed to shift flexibly to produce many different kinds of \textbf{chips} to suit demand. & computer chip, \textcolor{red}{fragment} & \textcolor{red}{cells}, \textcolor{teal}{processors}, \textcolor{teal}{screens} %\\ \hline
    %“Nathan,” said Nepthys, “I think that you just did something very stupid.” “I agree,” I \textbf{told} him. & speak to, reply to, warn, disclose, answer, express to, explain, chronicle, reckon, state, leave word, leak, fill in, proclaim, let know, inform, instruct, notify, communicate, utter, say to, let slip, confess, recite, authorize, mention, express, advise & informed, reassured, said \\ \hline
    \end{tabular}
%}
\label{tab:examples}
\end{table*}

As seen in Table \ref{tab:inspect}, both methods \say{miss} all of the gold labels very often, particularly in LS07 and Swords. Both methods, however, improve in the \textbf{T3C} metric from LS07 to CoInCo to Swords 5, implying that the benchmarks are in fact improvements to each other with regard to annotator agreement on the top-choice gold substitutes. The poor performance of both methods on the \textbf{MMP} metric also leads to interesting insights. While the task-based (Table \ref{tab:acc}) and similarity-based results (Table \ref{tab:CoInCo_cos}) demonstrate that contextual substitutions preserve semantic coherence and overall utility, the \textbf{MMP} scores imply that the model-predicted substitutes are not suitable at all. 

Table \ref{tab:examples} presents five randomly selected samples from the set of \textbf{MMP} results, i.e., \textsc{ConCat} output sets where none of the top-3 substitutes were found in the gold labels. Note that these cases imply the \textit{worst case} predictions of \textsc{ConCat}. As one can see, however, these \say{poor} \textsc{ConCat} substitutes often do contain either semantically and/or contextually relevant replacements, whereas these are simply not reflected in the gold labels. 
%For example, a replacement proposed by \textsc{ConCat} 
%in the third sentence of Table \ref{tab:examples} 
%is the lemma \textit{raged}, which is quite similar in meaning to the target. Similarly,
For example, \textit{complain} presents a good \textit{contextual} replacement, which is not covered in the benchmark annotation scheme.

These insights call into question whether any of the benchmark gold labels
%tested benchmarks, particularly their gold labels, do 
extend beyond synonym replacement to \textit{contextually relevant} replacements. Our analysis would imply this is not the case. Thus, we view that a closer investigation of the suggested alternatives be made for the LS field going forward.

\paragraph{User Preferences}
In the analysis of our survey results, one can see that although our method's substitutes were chosen the most, there is no clear majority. This leads us to hypothesize that preference for LS is highly subjective. Nevertheless, our results show that the contextually suitable replacements proposed by our model tend to be preferred. Taking a few examples of survey responses:

\begin{itemize}
\small
    \itemsep 0em
    \item[E1] Depending upon how many warrants and options are \textbf{exercised} prior to completion of the transaction...
    \begin{itemize}
        \itemsep 0em
        \item[\underline{Gold}]  use
        \item[\underline{Ours}]  execute
    \end{itemize}
\end{itemize}

\begin{itemize}
\small
    \itemsep 0em
    \item[E2] It \textbf{hissed} thoughtfully.
    \begin{itemize}
    \itemsep 0em
        \item[\underline{Gold}]  buzz, hoot, say
        \item[\underline{Ours}]  whisper, say, mutter 
    \end{itemize}
\end{itemize}

In example E1, \textbf{90.9\%} of respondents preferred our substitutes over the gold label; likewise, \textbf{86.4\%} preferred ours over the gold choices in E2. Ultimately, such findings suggest the need for a more in-depth study of the human perspective in LS, namely how LS methods reflect our way of thinking, in terms of synonymous versus contextual substitutions.

\section{CONCLUSION}
\label{sec:conclusion}
In this paper, we introduced \textsc{ConCat}, a simple and intuitive approach for English lexical substitution. The approach generates highly contextually fitting word substitutes while preserving the semantics of the surrounding sentence. We test our approach using established metrics on three standard LS benchmarks. Deeper analysis of the benchmark structure revealed certain weak points, which provided terrain for a critical discussion of current testing approaches in the LS community. For better insight into our approach’s usefulness, we conduct a qualitative survey where we assessed the user preferences.
%of candidate substitutes between the dataset’s ground truth, our approach, and a competing approach. 
The survey revealed users preferred \textsc{ConCat}’s substitutes more so than the competing approach and gold substitutes. Finally, we provided an analysis of how the performance of a text classification model changes when input instances have some of their words replaced with \textsc{ConCat} substitutes, showing that our approach is better at preserving the semantics needed for the classification performance than the competing approach. We hope \textsc{ConCat} will prove useful as a component of generative NLP tasks dealing with text simplification, stylistic transfer, or author anonymization. 

\section*{LIMITATIONS}
Despite showing impressive performance, our approach still falls short in some instances. There are cases when it generates incomplete words.
%and word pieces as its highest ranking candidates. 
Additionally, there are rare cases where filtering based on WordNet is too strict and removes word that could have been appropriate substitutes. Furthermore, our approach is only optimized for English The model also struggles with uncommon and rare words.
%, including foreign words inside of English sentences. 
% which was also influenced by the lack of datasets and resources for lexical substitution in other languages.

While we wished to provide a comprehensive comparison of our approach against competing approaches, we found it difficult to recreate them for a fair comparison. Even when public code repositories were provided, certain crucial files were missing or code dependencies were ill-defined. This includes the current state-of-the-art approaches ParaLS \citep{qiang-etal-2023-parals} and CILex \citep{seneviratne-etal-2022-cilex}, and their predecessor LexSubCon \citep{michalopoulos-etal-2022-lexsubcon}. Therefore, it is difficult to position our approach in the current research landscape. To account for this, we provided both a qualitative survey and NLP task performance experiments, which give more insight into the performance of \textsc{ConCat}.

% \section*{ACKNOWLEDGEMENTS}

\bibliographystyle{apalike}
{\small
\bibliography{example}}

\end{document}